\documentclass[article, 10pt, conference]{ieeeconf}      

\IEEEoverridecommandlockouts                              

\usepackage{spconf}

\usepackage{amssymb}
\usepackage{graphicx,epsfig,setspace,url,amsmath}
\usepackage{algorithm}
\usepackage[noend]{algpseudocode}
\usepackage{longtable}
\usepackage{array}
\usepackage{amsmath}
\usepackage{mathtools}
\usepackage{caption}
\usepackage{multirow}
\usepackage{relsize}
\usepackage{color}

\usepackage[]{subcaption}
\usepackage[pagebackref=false, breaklinks=true, letterpaper=true, colorlinks=false, citecolor=blue, bookmarks=false]{hyperref}

\usepackage{algorithm}
\usepackage[noend]{algpseudocode}

\title{Semi-supervised Multi-domain Multi-task Training \\ for Metastatic Colon Lymph Node Diagnosis From Abdominal CT}

 \name{Saskia Glaser$^{\dagger}$ \quad Gabriel Maicas$^{\dagger}$ \quad Sergei Bedrikovetski$^{\|}$ \quad   Tarik Sammour$^{\|\ddagger}$ \quad Gustavo Carneiro$^{\dagger}$ \sthanks{Supported by Australian Research Council through grant DP180103232 and by SA Health eHealth Innovation Grants Program (eIGP)}}
 \address {$^{\dagger}$ Australian Institute for Machine Learning, The University of Adelaide, Australia \\
 $^{\|}$Faculty of Health and Medical Science, School of Medicine, The University of Adelaide, Australia \\
 $^{\ddagger}$ Colorectal Unit, Department of Surgery, Royal Adelaide Hospital, Australia  }

\renewcommand{\thesection}{\arabic{section}} 
\renewcommand{\thesubsection}{\thesection.\arabic{subsection}}

\begin{document}

\maketitle
\thispagestyle{empty}
\pagestyle{empty}

\begin{abstract}
The diagnosis of the presence of metastatic lymph nodes from abdominal computed tomography (CT) scans is an essential task performed by radiologists to guide radiation and chemotherapy treatment.  
State-of-the-art deep learning classifiers trained for this task usually rely on a training set containing CT volumes and their respective image-level (i.e., global) annotation. However, the lack of annotations for the localisation of the regions of interest (ROIs) containing lymph nodes can limit  classification accuracy due to the small size of the relevant ROIs in this problem. 
The use of lymph node ROIs together with global annotations in a multi-task training process has the potential to improve classification accuracy, but the high cost involved in obtaining the ROI annotation for the same samples that have global annotations is a roadblock for this alternative. 
We address this limitation by introducing a new training strategy from two data sets: one containing the global annotations, and another (publicly available) containing only the lymph node ROI localisation.
We term our new strategy semi-supervised multi-domain multi-task training, where the goal is to improve the diagnosis accuracy on the globally annotated data set by incorporating the ROI annotations from a different domain. 
Using a private data set containing global annotations and a public data set containing lymph node ROI localisation, we show that our proposed training mechanism improves the area under the ROC curve for the classification task compared to several training method baselines.

\end{abstract}

\section{Introduction}
\label{sec:intro}

Colon cancer is one of the most frequently diagnosed cancers in the world, with nearly 1 million new cases and 551,269 deaths in 2018~\cite{bray2018global}. Patients with cancer have a high risk for nodal metastasis, therefore oncologically adequate surgery consists of segmental colectomy with lymph node dissection followed by adjuvant treatment~\cite{fernandez2019reliable}. However approximately 20\% to 30\% of patients develop another cancer months later, thus alternative treatment strategies including neoadjuvant therapy prior to surgery are currently being investigated~\cite{park2019potential}. Under these circumstances, accurate preoperative  metastatic lymph node diagnosis is crucial in determining the eligibility of neoadjuvant treatment and to avoid the over treatment of patients. However, diagnosing lymph nodes is a challenging task that is prone to inter-observer variability~\cite{choi2015accuracy}. As a result, computer-aided diagnosis systems are being designed to assist radiologists in staging lymph node metastasis on cross-sectional imaging.

State-of-the-art deep learning classification models~\cite{huang2017densely} could in principle produce relatively accurate diagnosis of metastatic lymph nodes. However, these models are trained with weakly annotated data sets (i.e. image-level, or global, labels), and fail to produce precise diagnosis in situations where the regions of interest (ROIs) that explain such diagnosis occupy a small portion of the input image, which is the case for metastatic lymph node diagnosis from abdominal CT scans.
The classification accuracy of models which perform diagnosis that depends on relatively small ROIs can be improved by incorporating localization information to the training process~\cite{lu2018identification,li2018thoracic,snaauw2019end}. The main goal is that the model learns to focus on the relevant ROIs to perform classification. For these models, the training process becomes a multi-task training, where the aims are to classify the entire image and to localize the ROIs in image. However, such training process is based on a strongly annotated training set that contains local ROI and global diagnosis annotations, which are costly and time consuming to obtain.

Aiming to reduce the burden of obtaining such ROI and global annotations for the single-domain multi-task training mentioned above, we hypothesize that global classification accuracy can be improved if we use separate data sets (i.e., multi-domain), each one containing one particular type of annotation (i.e., multi-task).  This setup introduces the new challenge of having a multi-domain multi-target training mechanism, where each domain (i.e., data set) is annotated for a different task -- this motivates us to propose a new semi-supervised learning approach that transfers the multi-task labels between multiple domains.

In this paper, we introduce a new semi-supervised training strategy to train multi-domain multi-task models where different domains have different types of annotations. Our strategy consists of jointly training a multi-task model with semi-supervised learning on the unlabelled task of the other domains. More precisely, we design a training strategy that improves image classification in one domain by including a ROI localization task from a different domain.
We evaluate our training method on the problem of metastatic colon lymph node diagnosis from abdominal CT, using a private data set that only contains global labels (RAH data set) and a public data set that only has lymph node ROI annotations (CI data set)~\cite{roth2014new}.
Results on the RAH data set show that our semi-supervised multi-domain multi-task training strategy outperforms other training strategies such as semi-supervised or multi-task learning without semi-supervision.

\section{Related Work}
\label{sec:lit_review}

Diagnosing metastatic lymph nodes from abdominal CT scans has traditionally been tackled with the design of hand-crafted features~\cite{huang2016development}. The main disadvantage of these methods lies in the sub-optimality of the features as there is no guarantee that these features are optimal to perform diagnosis, yielding relatively low accuracy scores. Feature sub-optimality has been addressed by the computer vision community by using deep learning classifiers that learn optimal features while being trained to perform diagnosis~\cite{huang2017densely}. However, the performance of such classifiers decreases in problems where the regions of interest in the image to be classified are relatively small.

The medical image analysis community focused on including the localization of regions of interest in the training process~\cite{lu2018identification,li2018thoracic,snaauw2019end} to increase classification performance, in a single-domain multi-task approach. The training process in such cases does not only optimise image classification, but also ROI localisation. Although successful, this increase in performance comes at the expense of strongly annotating training sets with ROIs and classification labels, a costly process for many medical image analysis problems.

Multi-task learning from multi-domain data~\cite{fourure2017multi} was proposed to learn from data sets that have been labelled for different but related tasks. The main aim is to improve the performance during inference on each of the labelled tasks in the data sets. However, these approaches require the multi-task annotations in all data sets to generalize well in each of the tasks across data sets, which is costly to be acquired.  Moreover, multi-domain multi-task methods do not use all the information available in the data sets as they do not exploit, during training, un-labelled data for a given task.

Semi-supervised learning has been proposed to incorporate un-annotated data for a given task into the learning process. Initially proposed for incorporating unlabelled data from the same task and domain~\cite{li2019signet,liu2019semi}, it has recently been successfully explored in a multi-task set-up~\cite{imran2019semi,chen2019multi}. 
Contrary to previous semi-supervised multi-task approaches~\cite{imran2019semi,chen2019multi}, we propose a training method that not only deals with tasks that have been labelled in the same domain, but also where each task is labelled in a different domain.


\section{Methods}
\label{sec:methodology}

\subsection{Data Sets}
\label{sec:method_data}

Let {\scriptsize ${\cal D}^{(1)} = \left \{ \left ( \mathbf{v}_i^{(1)}, y_i^{(1)}\right ) \right \}_{i \in \{1,...,|{\cal D}^{(1)}| \}}$} be a weakly annotated data set, where $\mathbf{v}_i^{(1)}:\Omega \rightarrow \mathbb R$ represents the abdominal CT scan of the $i^{th}$ patient in the data set ${\cal D}^{(1)}$ and $\Omega\in\mathbb {R}^3$ is the volume lattice, and $y_i^{(1)} \in \{ 0,1 \}$ is the scan-level label indicating the presence ($y^{(1)}_i = 1$) or absence ($y^{(1)}_i = 0$) of any metastatic lymph node in the CT scan of patient $i$. 
Let {\scriptsize ${\cal D}^{(2)} = \left \{ \left ( \mathbf{v}_j^{(2)}, \mathbf{s}_j^{(2)} \right ) \right \}_{j \in \{1,...,|{\cal D}^{(2)}| \}}$} be a strongly annotated data set where $\mathbf{v}_j^{(2)}:\Omega \rightarrow \mathbb R$ represents the abdominal CT scan of the $j^{th}$ patient in the data set ${\cal D}^{(2)}$, and $\mathbf{s}_j^{(2)}:\Omega \rightarrow \{0,1\}$ is the voxel-wise ROI annotation of the presence of a lymph node  in the CT scan of patient $j$. 
Assume that each data set comes from a different domain, implying that the data set distributions $(\mathbf{v},y) \sim P^{(1)}(\mathbf{v},y)$ (for $(\mathbf{v},y) \in \mathcal{D}^{(1)}$) and $(\mathbf{v}, \mathbf{s}) \sim P^{(2)}(\mathbf{v}, \mathbf{s})$ (for $(\mathbf{v}, \mathbf{s}) \in \mathcal{D}^{(2)}$) are different. Note that each data set is labelled for a different task: ${\cal D}^{(1)}$ for image classification and ${\cal D}^{(2)}$ for lymph node ROI detection. 
We split each data set into training, validation and testing sets in a patient-wise manner, forming the sets $\mathcal{T}_{train}^{(1)}$, $\mathcal{T}_{train}^{(2)}$,$\mathcal{T}_{val}^{(1)}$, $\mathcal{T}_{val}^{(2)}$, $\mathcal{T}_{test}^{(1)}$ and $\mathcal{T}_{test}^{(2)}$ for data sets ${\cal D}^{(1)}$ and ${\cal D}^{(2)}$.

\subsection{Semi-supervised Multi-domain Multi-task training}
\label{sec:method_model}

\begin{figure}[t]
\begin{center}
\includegraphics[width=0.45\textwidth]{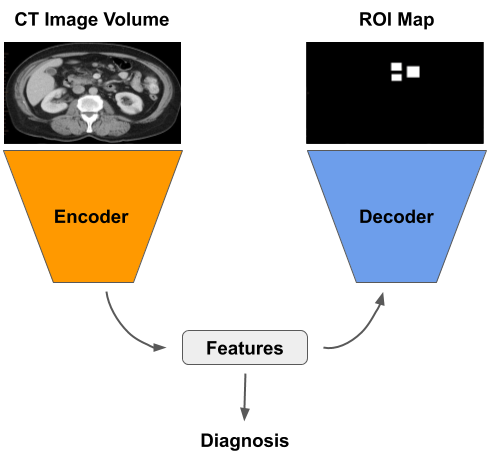} \end{center}
\vspace{-.1in} 
\caption{Architecture of our method. The model receives a scan $\mathbf{v}$ as input and the encoder builds its feature representation $\mathbf{o} = f_{\theta_e}(\mathbf{v})$, which is used by the classification branch $\sigma_{\theta_c}(\mathbf{o})$ and the ROI detection branch $g_{\theta_d}(\mathbf{o})$. The ROI binary map is obtained by thresholding the output from the ROI detection branch.}
\label{fig:intro}
\end{figure}

Our model is composed of three modules: an encoder $f_{\theta_{e}}$, a classification branch $\sigma_{\theta_{c}}$ and a detection branch $g_{\theta_{d}}$ with parameters $\theta_{e},\theta_{c},\theta_{d}$ respectively. 
The model receives as input a CT scan $\mathbf{v}$ and forms a feature embedding $\mathbf{o} = f_{\theta_{e}}(\mathbf{v})$ using the encoder. The embedding $\mathbf{o}$ is used as input by the classification and detection branches.
The classification branch returns $\tilde{y} = \sigma_{\theta_{c} }(\mathbf{o})$ representing the binary classification of the input scan $\mathbf{v}$. The detection branch produces an ROI map $\tilde{\mathbf{s}} = g_{\theta_{d}}(\mathbf{o})$ that estimates the probability that each voxel represents a lymph node. A binary ROI mask $\tilde{\mathbf{s}}_{\zeta}$ is generated by thresholding $\tilde{\mathbf{s}}(\omega) > \zeta$, for $\omega \in \Omega$. See Fig.~\ref{fig:intro} for a summary of the architecture of our method.

\textbf{The training} consists of two alternating stages: 1) label propagation between the two data sets sets~\cite{zhuѓ2002learning}, and 2) multi-domain multi-task training using the real and propagated labels. 
For label propagation between the two data sets~\cite{zhuѓ2002learning}, we replace the original $\mathcal{D}^{(1)}$ by $\mathcal{D}_{new}^{(1)} = \mathcal{D}^{(1)} \bigcup \widetilde{\mathcal{D}^{(1)}}$, with $\widetilde{\mathcal{D}^{(1)}} = \{ (\mathbf{v},\tilde{y}) | \mathbf{v} \in \mathcal{D}^{(2)}, \tilde{y} = \sigma_{\theta_c}(f_{\theta_e}(\mathbf{v}))  \}$ and the original $\mathcal{D}^{(2)}$ by $\mathcal{D}_{new}^{(2)} = \mathcal{D}^{(2)} \bigcup \widetilde{\mathcal{D}^{(2)}}$, with $\widetilde{\mathcal{D}^{(2)}} = \{ (\mathbf{v},\tilde{\mathbf{s}}) | \mathbf{v} \in \mathcal{D}^{(1)}, \tilde{\mathbf{s}} = g_{\theta_d}(f_{\theta_e}(\mathbf{v})) \}$.

For the multi-domain multi-task stage, we jointly minimise the classification loss with 
\begin{equation}
\theta_c^*,\theta_e^* = \arg \max_{\theta_c,\theta_e} \mathbb E_{P^{(1)}(\mathbf{v},y)} [ \ell_C( \sigma_{\theta_c} ( f_{\theta_e}(\mathbf{v}) ),y )],
\label{eq:clf_loss}
\end{equation}
where $\ell_C$ denotes the binary cross entropy loss, and the ROI detection loss with
 \begin{equation}
 \theta_d^*,\theta_e^* = \arg \max_{\theta_d,\theta_e} \mathbb E_{P^{(2)}(\mathbf{v},\mathbf{s})} [ \ell_D( g_{\theta_d} ( f_{\theta_e}(\mathbf{v}) ), \mathbf{s} ) \} ],
 \label{eq:det_loss}
\end{equation}
where $\ell_D$ denotes a loss that sums the voxel-wise cross entropy loss and the Dice loss~\cite{milletari2016v}.  
We only use the subsets $\mathcal{T}_{train}^{(1)}$, $\mathcal{T}_{train}^{(2)}$,$\mathcal{T}_{val}^{(1)}$, $\mathcal{T}_{val}^{(2)}$ from $\mathcal{D}^{(1)}$ and $\mathcal{D}^{(2)}$ during the training process.

\textbf{The inference} consists of feeding the model with an input scan $\mathbf{v}$ that is encoded to obtain the feature embedding and forwarded through the classification branch $\sigma_{\theta_c^*}(f_{\theta_e^*}(\mathbf{v}))$ to yield the probability of metastatic lymph nodes.

\section{Experiments and Results}
\label{sec:Experiments} 

We assess our proposed method on the problem of diagnosing the presence of metastatic lymph nodes from abdominal CT scans. For the weakly labelled data set ${\cal D}^{(1)}$ we use a private data set from the Royal Adelaide Hospital (RAH data set) 
that contains 123 scans from 123 patients. Weak labels indicating the presence of any metastatic lymph node are obtained from pathology reports and clinical notes. There are 57 scans labelled with the presence of metastatic lymph nodes and 66 scans labelled with the absence of metastatic lymph nodes.
Note that no lymph node localization is provided in this data set. 
The data set ${\cal D}^{(2)}$ that provides lymph node localization information is publicly available~\cite{roth2014new} (CI data set). There are 595 ROI localisation annotations of lymph nodes from 85 scans of 85 patients. Note that no scan-level label about metastatic lymph node is provided with this data set. 
The RAH data set is randomly divided into training (90 patients, 43 of them with metastatic lymph nodes and 47 without), validation (10 patients, 4 with metastatic lymph nodes, 6 without), and testing (23 patients, 10 with metastatic lymph nodes, 13 without). 
The CI data set is randomly divided into training (62 patients with 439 annotated lymph nodes), validation (7 patients with 54 annotated lymph nodes), and testing (16 patients with 102 lymph nodes). We train our method with the training set from RAH and the entire CI data set. We utilize the validation set of the RAH data set to perform model selection and we report classification results on the testing set of the RAH data set. 
We pre-process each image in both data sets by subtracting the mean and dividing by the standard deviation of the training set of the corresponding data set. 

The encoder $f_{\theta_{e}}(.)$ is represented by a 3-D Densenet~\cite{huang2017densely} consisting of 5 dense blocks. The input volume $\mathbf{v}$ size is $512\times512\times256$ and the feature vector embedding $\mathbf{o} = f_{\theta_{e}}(\mathbf{v})$ is of size $16\times16\times8$.
The classifier $\sigma_{\theta_c}(.)$ is composed of two fully connected layers and outputs the probability of presence of metastatic lymph nodes in $\mathbf{v}$. 
The decoder $g_{\theta_{d}}(.)$ is also composed of 5 dense blocks, and outputs an ROI map $\tilde{\mathbf{s}}$ of the same size as the input scan and is thresholded at $\zeta = 0.8$ to obtain $\tilde{\mathbf{s}}_{\zeta}$. 
The training process runs for 500 epochs and optimizes the parameters $\theta_{e},\theta_{c},\theta_{d}$ with Adam optimizer (learning rate of 0.05).

We evaluate our proposed semi-supervised multi-domain multi-task training method with the metastic lymph node classification performance, measured with the area under the ROC curve (AUC). We compare our proposed training strategy against other training procedures: 1) \textbf{supervised baseline:} a DenseNet~\cite{huang2017densely} classifier trained on the RAH data set and composed of the encoder and classification branch;
2) \textbf{semi-supervised:} the same DenseNet classifier from (1), trained with the RAH training set and including the CI data set trained with propagated classification labels; and 3) \textbf{supervised multi-domain multi-task:} the proposed architecture trained with the RAH and CI data sets, where the encoder and classifier are trained with the RAH data set and the encoder and detector branches are trained with CI data set. 
We present quantitative classification results in terms of AUC on Table~\ref{tab:Baselines}.

\begin{table}[]
    \centering
    \scalebox{0.86}{
        \begin{tabular}{|l|c|c|}
            \hline
            \textbf{Training Method for the classifier} & \textbf{Data Sets} & \textbf{AUC} \\
            \hline
            \hline
            Supervised Baseline & RAH & 0.81  \\
            \hline
            Semi-Supervised & RAH + CI & 0.84  \\
            \hline
            Supervised Multi-domain Multi-Task    &RAH + CI & 0.82       \\
            \hline
            \textbf{Semi-Supervised Multi-domain Multi-Task} & RAH + CI &  \textbf{0.86}\\
            \hline
        \end{tabular}
    }
    \caption{Classification AUC on the RAH Test Set obtained by the classifier trained with our proposed training strategy and baseline methods.}
    \label{tab:Baselines}
\end{table}

\begin{figure}[t]
    \centering
    \begin{subfigure}[t]{0.45\linewidth}
        \centering
        \includegraphics[width = \linewidth]{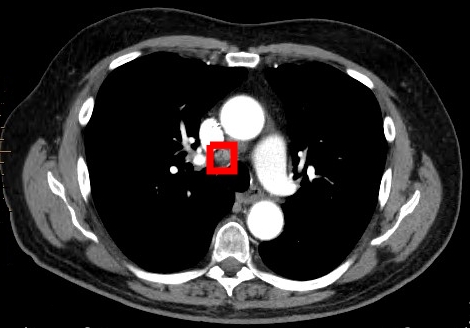}
        \caption{}
        \label{img1}
    \end{subfigure}
    \begin{subfigure}[t]{0.45\linewidth}
        \centering
        \includegraphics[width = \linewidth]{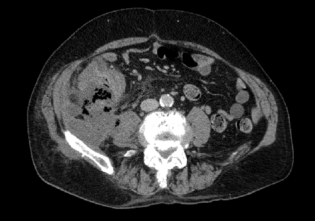}
        \caption{}
        \label{img2}
    \end{subfigure}
    \caption{Visual results produced by our proposed method on the RAH data set. Image~\ref{img1} shows the positive classification of an scan containing a metastatic lymph node (marked in red). Image~\ref{img2} contains the negative classification of an image with non-metastatic lymph nodes.}
    \label{fig:res}
\end{figure}


 \section{Discussion and Conclusion}
\label{sec:dic}

The experimental results presented in Table~\ref{tab:Baselines} show that our proposed training method outperforms several baseline training strategies for metastatic lymph node diagnosis from abdominal CT scans. As explained in Sec.~\ref{sec:intro}, the baseline classifier achieves the lowest AUC score, which can be in part due to the lack of lymph node localization. Interestingly, propagating the classification labels from RAH to CI data set (without including any localization in the training process) yielded better results for classification than the supervised multi-domain multi-task baseline. We believe this is due to difficulty of integrating multiple tasks from different domains into the training process without any semi-supervision.
Finally, our proposed semi-supervised multi-domain multi-task training outperformed all baseline methods. As we hypothesized in Sec.~\ref{sec:intro}, jointly semi-supervising each of the labelled tasks from a different domain results in a more accurate model, probably due to the extra-supervision that can facilitate the addition of data from a different domain. 

In conclusion, we proposed a new 
semi-supervised multi-domain multi-task training, where we semi-supervised the detection task of the data set containing global annotations and the classification task on the data set containing ROI annotations. 
Results on diagnosing the presence of metastatic lymph nodes from CT scans showed that our method successfully incorporates lymph node localization information from a different domain to improve classification results in the original domain. We leave for future work the evaluation of the lymph node ROI localization task in each data set. 

\bibliographystyle{IEEEbib}
\bibliography{bibli.bib}

\end{document}